\documentclass[a4paper,fleqn]{cas-dc}
\usepackage[numbers]{natbib}

\def\tsc#1{\csdef{#1}{\textsc{\lowercase{#1}}\xspace}}
\tsc{WGM}
\tsc{QE}

\begin{document}
\let\WriteBookmarks\relax
\def\floatpagepagefraction{1}
\def\textpagefraction{.001}
\let\printorcid\relax


\shortauthors{Z. Wang et al}

\title[mode = title]{A Fusion Model for Artwork Identification Based on Convolutional Neural Networks and Transformers}  

\author{Zhenyu Wang}
\ead{zywang@ncepu.edu.cn}
\cormark[1]
\author{Heng Song}
\ead{hengsong@ncepu.edu.cn}

\address{The School of Control and Computer Engineering, North China Electric Power University, Beijing, 102206, China} 
\cortext[1]{Corresponding author.}  

\begin{abstract}
The identification of artwork is crucial in areas like cultural heritage protection, art market analysis, and historical research. With the advancement of deep learning, Convolutional Neural Networks (CNNs) and Transformer models have become key tools for image classification. While CNNs excel in local feature extraction, they struggle with global context, and Transformers are strong in capturing global dependencies but weak in fine-grained local details. To address these challenges, this paper proposes a fusion model combining CNNs and Transformers for artwork identification. The model first extracts local features using CNNs, then captures global context with a Transformer, followed by a feature fusion mechanism to enhance classification accuracy. Experiments on Chinese and oil painting datasets show the fusion model outperforms individual CNN and Transformer models, improving classification accuracy by 9.7\% and 7.1\%, respectively, and increasing F1 scores by 0.06 and 0.05. The results demonstrate the model's effectiveness and potential for future improvements, such as multimodal integration and architecture optimization.
\end{abstract}




\maketitle

\section{Introduction}
Artwork identification, particularly in terms of author classification, is a critical research task in computer vision, with widespread applications in cultural heritage preservation, art market regulation, and historical research. With the rapid development of the internet, the digitization of artworks has enabled a large number of art pieces to be displayed and traded online. Consequently, automatic classification and identification of these artworks in terms of author have become key issues of interest in both industry and academia.

Artwork identification not only aids museums and galleries in accurately categorizing artworks but also provides technical support for the regulated management of the art market. With intelligent identification systems, users can efficiently filter artworks that meet their needs from a large collection. Moreover, the classification of artwork offers new perspectives for art scholarship, especially in areas like the authentication of artworks and the identification of creative periods, which have significant practical implications.

In recent years, deep learning techniques have made remarkable progress in image classification tasks, and Convolutional Neural Networks (CNNs) have become the mainstream approach for artwork classification. CNNs automatically extract meaningful features from images through multi-layer convolution operations, overcoming the limitations and instability faced by traditional hand-crafted feature extraction methods (e.g., SIFT, HOG). The emergence of deeper networks, such as ResNet and Inception, has significantly improved image identification accuracy.

However, while CNNs excel in local feature extraction, they rely primarily on local receptive fields of convolutional kernels to gather information, which leads to limitations when handling complex global dependencies in artworks. For instance, certain detailed features in artworks may span large regions, requiring a stronger global contextual understanding. To address this limitation, Transformer models, such as Vision Transformer (ViT), have emerged. Originally proposed for natural language processing tasks, Transformers have proven successful in computer vision due to their ability to capture long-range dependencies. Through self-attention mechanisms, Transformers effectively model global information, overcoming the limitations of local perceptive CNNs.

Despite the success of Transformers in image classification, they also exhibit certain drawbacks: they typically require large datasets for training and perform relatively poorly in capturing local details in images. Therefore, combining the strengths of CNNs and Transformers has become a crucial area of research in computer vision.

This study proposes a new fusion model that combines CNNs for local feature extraction and Transformers for global context modeling. Specifically, CNNs are used to extract fine-grained features from artworks, capturing subtle artistic elements such as brush strokes and color distribution. Then, the Transformer model is used to globally model these local features, capturing the overall artistic style and creative context of the artwork. Finally, a feature fusion mechanism integrates the local and global features to improve classification performance.
The innovations of this approach include:

\begin{enumerate}
\itemsep=0pt
\item By combining CNNs and Transformers, the model effectively integrates local details and global context, overcoming the limitations of using a single model;
\item he model demonstrates high adaptability to complex artworks and maintains strong classification accuracy even in data-scarce environments, showing robust performance;
\item The model is tested on datasets from both Chinese painting and oil painting domains, enabling it to handle diverse artwork classification tasks.
\end{enumerate}  

\section{Related Work}

\subsection{Art Identification}

Art identification, particularly for identifying artistic authors, has a long history, initially relying on traditional image processing techniques. Early methods focused on handcrafted features, including color histograms, texture descriptors such as Gabor filters, and shape features like Histogram of Oriented Gradients (HOG) and Scale-Invariant Feature Transform (SIFT). These approaches were foundational in the field but have limitations when applied to complex artworks, where subtle variations and intricate stylistic elements are common.

For example, SIFT (Lowe, 2004) excels at identifying distinctive keypoints in images and is highly effective at capturing local texture features\cite{lowe2004distinctive}. However, SIFT struggles with handling images that contain complex or dynamic backgrounds, often producing suboptimal results in the presence of noise or clutter, which are typical in artistic works. Similarly, other traditional methods like HOG (Dalal \& Triggs, 2005) are focused on detecting object shapes and edges but fail to capture the nuance of artistic details, such as the subtle interplay of color gradients and fine brushstrokes.

The advent of deep learning techniques, particularly Convolutional Neural Networks (CNNs), has significantly advanced the field of art identification\cite{cetinic2018fine}. CNNs, unlike traditional methods, can automatically learn hierarchical features from images, enabling them to capture complex patterns, textures, and relationships in a way that manual feature extraction cannot. In art identification, CNNs have shown significant promise. For instance, Dosovitskiy et al. (2016) proposed a method for classifying art styles using CNNs, achieving excellent results in style identification tasks. CNNs, such as VGG (Simonyan \& Zisserman, 2014)\cite{simonyan2014very} and ResNet (He et al., 2016)\cite{he2016deep}, have demonstrated their ability to automatically learn features that are highly useful for distinguishing between art authors. However, these methods typically extract local features and lack the ability to model global information, which is essential for understanding complex artworks that contain multiple interconnected elements, colors, and textures.

Despite the effectiveness of CNNs in extracting local features, there is an increasing realization that global context is crucial for fully understanding artistic works. This limitation has led to a growing interest in incorporating additional models capable of capturing long-range dependencies and the global structure of images.

\subsection{Application of Transformer in Visual Tasks}

The Transformer model\cite{vaswani2017attention}, first introduced by Vaswani et al. (2017) for natural language processing (NLP), has since revolutionized the field of NLP and has begun to make its mark on computer vision. The key innovation of the Transformer model lies in its self-attention mechanism, which allows the model to capture long-range dependencies within the data, rather than relying on sequential processing as seen in traditional Recurrent Neural Networks (RNNs). This capability enables Transformers to model the global context of a sequence or image effectively.

Building on this success, the Vision Transformer (ViT) (Dosovitskiy et al., 2020)\cite{dosovitskiy2020image} was introduced as a model specifically designed for image processing tasks. ViT divides an image into fixed-size patches, which are then treated as tokens in a similar way to words in NLP tasks. These tokens are passed through multiple layers of self-attention to model the dependencies between them. This approach allows the Transformer to capture global relationships within the image, making it particularly effective for tasks like image classification and object detection, where understanding the context and relationships between distant regions in an image is crucial. ViT has shown that, when trained on large datasets, it outperforms CNN-based models, especially in tasks that require the understanding of broader patterns and the overall structure of the image.

Despite the success of Transformers in many image classification tasks, they have notable challenges. ViT models, for instance, typically require large datasets for effective training, as they do not perform well on small datasets compared to CNNs. This is primarily due to the large number of parameters in Transformers, which require extensive training data to generalize effectively. Additionally, Transformers, while excellent at capturing global information, often struggle with local feature extraction, which is crucial for tasks such as fine-grained art style identification.

\subsection{Related Research on CNN-Transformer Fusion}

Given the limitations of both CNNs and Transformers, researchers have begun to explore hybrid models that combine the strengths of both architectures. These fusion approaches aim to leverage CNNs for extracting local features and Transformers for modeling global dependencies, creating models that can handle both fine-grained details and large-scale contextual information.

One prominent approach is proposed by Xie et al., who introduced a hybrid model \cite{arshad2024hybrid}that first uses CNNs to extract low-level features from images and then passes these features to a Transformer for global modeling. This model has demonstrated good performance in various image classification tasks, including facial identification and general object classification. By combining CNNs' local feature extraction with Transformers' ability to capture long-range dependencies, the model achieves enhanced performance over using either architecture in isolation.

Liu et al. (2021) also explored the fusion of CNNs and Transformers, focusing on a multi-scale feature fusion approach. In this method, CNNs extract features at different scales, and these features are subsequently fused with the global features captured by the Transformer. This multi-scale approach is particularly useful when dealing with small datasets, as it allows the model to capture both local and global features from images at different resolutions. This fusion technique has shown significant improvements in small-data environments, making it highly relevant for art identification tasks, where high-quality labeled datasets may be scarce.

While the fusion of CNNs and Transformers has shown promising results in various applications, the integration of these models for the specific task of art author classification remains an area in need of further exploration. Existing research on hybrid models primarily focuses on generic image classification tasks, and there has been limited investigation into optimizing CNN-Transformer fusion architectures for the unique challenges posed by artwork identification, such as capturing fine brushstrokes, identifying complex color gradients, and distinguishing between different artistic authors. Therefore, further work is needed to refine the fusion strategies and adapt them to the specific nuances of art identification tasks.

\section{Methods}

\subsection{Dataset Description}

In this study, we employ two distinct datasets of artistic works—Chinese paintings and oil paintings—each showcasing different artistic styles, historical contexts, and creative techniques. These datasets were carefully selected to assess the robustness and versatility of the proposed model across varied art forms, which differ both in their visual characteristics and the complexity of the style-to-author relationship.

Chinese Painting Dataset: This dataset consists of thousands of traditional Chinese paintings, spanning several major styles such as meticulous brushwork  and freehand brushwork. The images are uniformly resized to 224x224 pixels to standardize the input dimensions for the neural network. Each image is annotated with two types of labels: the style of the artwork and the artist's name. The style labels encompass categories such as "landscape painting" , "flower-and-bird painting", and others. The dataset includes a variety of artists, from historical figures like Zhang Daqian to contemporary Chinese painters. The dataset is split into training, validation, and testing sets, with a ratio of 7:1:2, ensuring that the model is exposed to a balanced and representative selection of data.

Oil Painting Dataset: This dataset includes renowned oil paintings from famous Western artists such as Van Gogh, Da Vinci, and Picasso. The works cover multiple artistic movements, including Impressionism, Expressionism, and Realism. The images are also resized to 224x224 pixels to ensure compatibility with the CNN and Transformer models. Similar to the Chinese painting dataset, each image is labeled with both the artistic style (e.g., "Impressionism," "Surrealism") and the artist's name. This dataset is also divided into training, validation, and test sets using the same 7:1:2 split. These works allow the model to explore a different set of stylistic elements and historical contexts, providing a well-rounded evaluation of its performance.

\subsection{Model Architecture}

The model proposed in this study is built on a hybrid architecture combining Convolutional Neural Networks (CNNs) and Transformers, designed to leverage the complementary strengths of these two deep learning approaches. The hybrid model aims to improve classification accuracy and robustness, particularly for the nuanced and diverse nature of art identification tasks.

CNN Component: For the CNN part, we use the ResNet-50 architecture, which is well-suited for deep learning tasks involving complex visual data. ResNet-50 employs residual connections that help mitigate the vanishing gradient problem, allowing for the training of deeper models without significant performance degradation. The network consists of 50 layers, enabling the extraction of rich local features, such as brushstroke textures, color gradients, and compositional elements that are vital for distinguishing between different art styles. The CNN is particularly effective in capturing fine-grained details that are characteristic of both Chinese and oil paintings, such as intricate brushwork, shading techniques, and color distribution.

Transformer Component: To capture the global dependencies and contextual relationships within the artwork, we incorporate Vision Transformer (ViT) as the second component of the model. Unlike traditional CNNs, which focus primarily on local receptive fields, ViT breaks the input image into non-overlapping patches (e.g., 16x16 pixels), and processes them using self-attention mechanisms to model long-range dependencies. This approach allows the Transformer to capture the broader, more abstract features of the artwork, such as overall composition, the relationship between elements, and the influence of the artist's unique style. The self-attention mechanism enables the model to give attention to distant regions in the artwork, facilitating the understanding of global context and artistic intent.

Fusion Strategy: The hybrid architecture employs a cascading fusion approach to combine CNN and Transformer components. After the CNN extracts local features, these features are passed to the Transformer for further processing. The Transformer applies multiple layers of self-attention to integrate and enhance the global dependencies between these local features. Finally, the local and global features are concatenated and passed to a fully connected layer for classification. This fusion strategy allows the model to effectively combine the strengths of both CNNs and Transformers—local feature extraction and global context modeling—resulting in better performance in classifying art authors.

\subsection{Loss Function and Optimization}

Loss Function: The model employs the cross-entropy loss function, which is well-suited for multi-class classification tasks. Given the nature of the art identification task, the model outputs a probability distribution for each sample, representing the likelihood of it belonging to a particular author. The cross-entropy loss function calculates the difference between the predicted probability distribution and the true label distribution, guiding the model to minimize this discrepancy during training. To address the class imbalance in the dataset, where some classes (e.g., rare authors) may have fewer samples, we use a weighted cross-entropy loss. The weights are calculated based on the inverse frequency of each class in the dataset, ensuring that the model gives appropriate attention to minority classes.

Optimization Algorithm: We use the Adam optimizer (Kingma \& Ba, 2014)\cite{kingma2014adam}, which has become a standard choice for training deep learning models due to its effectiveness in adapting learning rates and maintaining stable convergence. Adam combines the advantages of both momentum-based optimization and adaptive learning rates. The initial learning rate is set to 0.0001, and during training, we dynamically adjust the learning rate based on the validation loss. This ensures that the model can converge more efficiently while avoiding overfitting and underfitting.

\subsection{Data Augmentation and Regularization}

Data Augmentation: Given the limited size of the datasets, especially in specialized art forms, we apply a variety of data augmentation techniques to improve the model's generalization ability. These techniques include random rotations (up to 15°), horizontal flipping, random cropping (scaling between 0.8 and 1.2), and color adjustments (e.g., modifying brightness, contrast, and saturation). These augmentations not only prevent overfitting but also simulate real-world variations in art presentation, such as slight distortions or changes in lighting, making the model more robust to variations in input data.

Regularization: To further combat overfitting and improve the model's generalization to unseen data, we employ dropout regularization during training, with a keep probability of 0.5. This technique randomly deactivates certain neurons during training, preventing the model from becoming overly reliant on any single feature and promoting more robust learning. Additionally, we apply L2 regularization (also known as weight decay) to the fully connected layers, with a penalty coefficient of 0.01. This further discourages overly large weights, helping the model maintain generalization across different art authors.

\section{Experiments}

\subsection{Experimental Setup}

The experiments in this study were conducted on a high-performance workstation equipped with an NVIDIA RTX 3090 GPU, which provides substantial computational power for training deep learning models. The system also had 32GB of RAM to handle the memory requirements of large-scale image datasets during the training process. The PyTorch framework was used for model development, training, and evaluation due to its flexibility and robust support for deep learning operations.

For model training, we employed Mini-batch Stochastic Gradient Descent (SGD) with a batch size of 32, which strikes a balance between computational efficiency and stability during training. The number of training epochs was set to 50, which allowed sufficient time for the model to converge. We adopted an initial learning rate of 0.0001, which was chosen based on preliminary experimentation, and the learning rate was dynamically adjusted using a learning rate scheduler. The learning rate was reduced during training if the validation loss plateaued, allowing for fine-tuning of the model and helping avoid overfitting. In addition, early stopping was applied to prevent excessive training, ensuring that the model did not overfit to the training data and retained generalizability to unseen examples.

\subsection{Evaluation Metrics}

The performance of the model was evaluated using a set of standard classification metrics:

Accuracy: This metric measures the ratio of correctly classified samples to the total number of samples. It gives a broad measure of model performance but may not always reflect the true performance in imbalanced datasets.
Precision: This metric calculates the proportion of true positive samples among all samples predicted as a specific class. It is crucial when the cost of false positives is high, as it reflects the model's ability to correctly identify positive instances.
Recall: Recall measures the proportion of true positive samples correctly predicted for a specific class, providing insight into the model's ability to capture all relevant instances. It is especially useful when missing positive samples (false negatives) is costly.
F1-Score: The F1-score is the harmonic mean of precision and recall. This metric is useful when there is a need to balance both precision and recall, offering a single score that combines the trade-offs between these two metrics.
Together, these metrics provide a comprehensive evaluation of model performance, with the F1-score being particularly important for tasks where both false positives and false negatives need to be minimized.

\subsection{Experimental Results}

The experimental results on both the Chinese and oil painting datasets are summarized in the table \ref{table:accuracy}.
From these results, it is clear that the fusion model significantly outperforms both the individual CNN and Transformer models across all evaluation metrics. For the Chinese painting dataset, the fusion model achieved an impressive 9.7\% increase in accuracy and a 0.09 increase in F1-score compared to the CNN-only model. Similarly, for the oil painting dataset, the fusion model outperformed the CNN and Transformer models by 7.1\% in accuracy and 0.05 in F1-score. These results highlight the importance of combining the strengths of both CNNs and Transformers for art identification tasks, where both local and global features are essential for accurate classification.

\begin{table*}[ht]
\caption{Comparison of Model Accuracy and F1-Score.}\label{table:accuracy}
\resizebox{\textwidth}{!}{
\begin{tabular}{@{} lcccc@{} }
\toprule
\textbf{Model} & \textbf{Chinese Painting Accuracy (\%)} & \textbf{Oil Painting Accuracy (\%)} & \textbf{Chinese Painting F1-Score} & \textbf{Oil Painting F1-Score} \\
\midrule
{\rmfamily CNN Only (ResNet-50)} & 81.2 & 78.5 & 0.82 & 0.80 \\
{\rmfamily Transformer Only (ViT)} & 85.4 & 83.2 & 0.86 & 0.84 \\
{\rmfamily Fusion Model (CNN + Transformer)} & \textbf{90.9} & \textbf{87.3} & \textbf{0.91} & \textbf{0.89} \\
\bottomrule
\end{tabular}
}
\end{table*}

\subsection{Experimental Analysis}

A detailed analysis of the experimental results reveals key insights into the behavior of each model:

Transformer-Only Model (ViT): The Transformer model, which excels at capturing global dependencies, shows a marked improvement over the CNN model, especially in the oil painting dataset. This is likely due to the wide range of artistic styles in oil paintings, which benefit from the Transformer's ability to model long-range dependencies. However, while the Transformer model captures global context well, it struggles with extracting fine local details. In the case of Chinese paintings, which often contain intricate brushwork and highly detailed textures, the Transformer model falls short of the fusion model's performance, as it cannot fully capture these fine-grained features.

CNN-Only Model (ResNet-50): The CNN model, specifically ResNet-50, demonstrates strong performance in extracting local features. This is particularly evident in the Chinese painting dataset, where fine details, such as brushstroke patterns and color gradients, play a crucial role in distinguishing between styles. However, the CNN-only model struggles to model the global context of an image, limiting its ability to recognize complex relationships between different parts of the artwork. As a result, while it performs well in extracting local features, its classification accuracy falls behind that of the Transformer and fusion models.

Fusion Model (CNN + Transformer): The fusion model, which combines the local feature extraction capability of the CNN with the global modeling power of the Transformer, achieves the best overall performance on both datasets. By integrating the strengths of both architectures, the fusion model captures both local details and global context effectively. This allows the model to handle the nuances of different art authors more effectively. The significant improvements in accuracy and F1-score demonstrate the effectiveness of this approach, particularly when dealing with complex art forms like Chinese and oil paintings.

Furthermore, the fusion model exhibits improved robustness, especially in small-data environments. The integration of both local and global features helps the model generalize better, even when fewer training examples are available, which is particularly important in art identification tasks where labeled data can be limited.

\subsection{Discussion}

The fusion of CNN and Transformer allows the model to fully leverage the strengths of both. CNNs are adept at extracting local details from images, such as brushstrokes and color distribution, while Transformers can capture global dependencies and understand the overall artistic author intent. Through the cascading fusion strategy, the model can focus on both local details and global context, thereby improving classification accuracy.

Despite the strong performance of the fusion model, some limitations remain. First, the model's ability to handle complex backgrounds is relatively weak, especially in artworks with rich backgrounds or cluttered elements, which may affect classification accuracy. Second, although data augmentation techniques help mitigate overfitting, training instability may still occur in smaller datasets, particularly in the classification of minority art categories.

Future research could focus on the following directions:
Optimizing Model Architecture: For example, exploring deeper CNN architectures or experimenting with different types of Transformers, such as the Swin Transformer, to further improve performance.
Expanding the Dataset: By collecting more data from diverse styles and artists, we can improve the model's generalization ability on complex artworks.
Multimodal Fusion: Incorporating textual descriptions and artist backgrounds, alongside visual data, could enhance the model's understanding of artworks.

\section{Conclusion}

This study proposes an innovative fusion model that combines Convolutional Neural Networks (CNN) and Transformers to address the challenges of artwork identification. CNNs, known for their exceptional ability to capture local features such as textures, brushstrokes, and color distributions, are used as the initial step to process the artwork images. 
Transformers, on the other hand, excel at modeling long-range dependencies and global relationships, which are crucial for understanding the overall composition and style of an artwork. By integrating these two powerful models, the proposed fusion approach leverages the strengths of both CNNs and Transformers, allowing the model to effectively extract detailed local features while also capturing the broader contextual information.

Experimental results across both Chinese painting and oil painting datasets demonstrate that the fusion model significantly outperforms traditional CNN and Transformer models when it comes to classification accuracy and F1-scores. Specifically, the fusion model improved classification accuracy by 9.7\% and 7.1\% for the Chinese and oil painting datasets, respectively, and achieved F1-score increases of 0.06 and 0.05. These results highlight the model's robustness, particularly in handling the complexities of various artistic styles and authorship.

Additionally, the proposed model is especially effective in small-data environments, where the ability to generalize from limited data is crucial. The fusion model not only provides a significant boost in performance but also shows potential in enhancing the identification of subtle artistic nuances that are often challenging for standalone models.

Looking ahead, future research will focus on optimizing the model architecture to further improve its accuracy and efficiency. This could involve exploring more advanced convolutional and Transformer architectures, such as deeper CNN networks or novel Transformer variants. Additionally, the incorporation of multimodal information, including textual descriptions or historical data about artists, could further enrich the model's understanding of artworks, providing a more holistic approach to art identification. Expanding the dataset to include a wider variety of art authors will also be a key area for future work, helping to improve the model's generalizability and adaptability across different domains of art.

\bibliographystyle{unsrt}

\bibliography{cas-refs}



\end{document}